\documentclass[runningheads]{llncs}

\usepackage{eccv}

\usepackage{eccvabbrv}

\usepackage{graphicx}
\usepackage{booktabs}
\usepackage{multirow}
\usepackage[table]{xcolor}

\usepackage[accsupp]{axessibility}  

\usepackage{siunitx}

\usepackage{hyperref}

\usepackage{orcidlink}

\begin{document}

\title{\textsc{VINO}: Video-driven Invariance for Non-contextual Objects via Structural Prior Guided De-contextualization} 

\titlerunning{\textsc{VINO}}

\author{Seul-Ki Yeom\orcidlink{0000-0002-3127-7219} \and
Marcel Simon\orcidlink{1111-2222-3333-4444} \and
Eunbin Lee\orcidlink{2222-3333-4444-5555} \and 
Tae-Ho Kim\orcidlink{2222-3333-4444-5555}} 


\institute{Nota AI GmbH \\
\email{\{skyeom,marcel.simon,eunbin.lee,thkim\}@nota.ai}}

\maketitle

\begin{abstract}
Self-supervised learning (SSL) has made rapid progress, yet learned features often over-rely on contextual shortcuts—background textures and co-occurrence statistics. While video provides rich temporal variation, dense \emph{in-the-wild} streams with strong ego-motion create a co-occurrence trap: foreground objects and background context move coherently, encouraging representations to collapse into scene encoders.

To address this, we propose \textsc{VINO} (Video-driven Invariance for Non-Contextual Objects), a teacher–student framework that learns robust image encoders from dense video by imposing a structural information bottleneck. Using a class-agnostic structural prior solely to generate views—not as semantic pseudo-labels—\textsc{VINO} forms an asymmetric distillation problem. The teacher predicts from a foreground-union view with the background suppressed, while the student observes object-conditioned scene views that retain surrounding context but remove competing instances. Matching these targets via masked distillation makes background cues unreliable, pushing the representation toward object-centric invariances. We further enforce temporal object permanence via teacher-anchored cross-time distillation over track-matched objects, and stabilize part-to-whole consistency with mask-guided local views.

Through attention visualization and unsupervised object discovery on PASCAL VOC, we demonstrate that \textsc{VINO} effectively disentangles foreground from background. Pretrained on the dense Walking Tours Venice video, \textsc{VINO} achieves 34.8 CorLoc, yielding highly focused, shape-biased representations that substantially outperform prior dense-video and motion-guided SSL baselines.
\end{abstract}

\begin{figure}[t!]
  \centering
  \includegraphics[width=1.0\columnwidth]{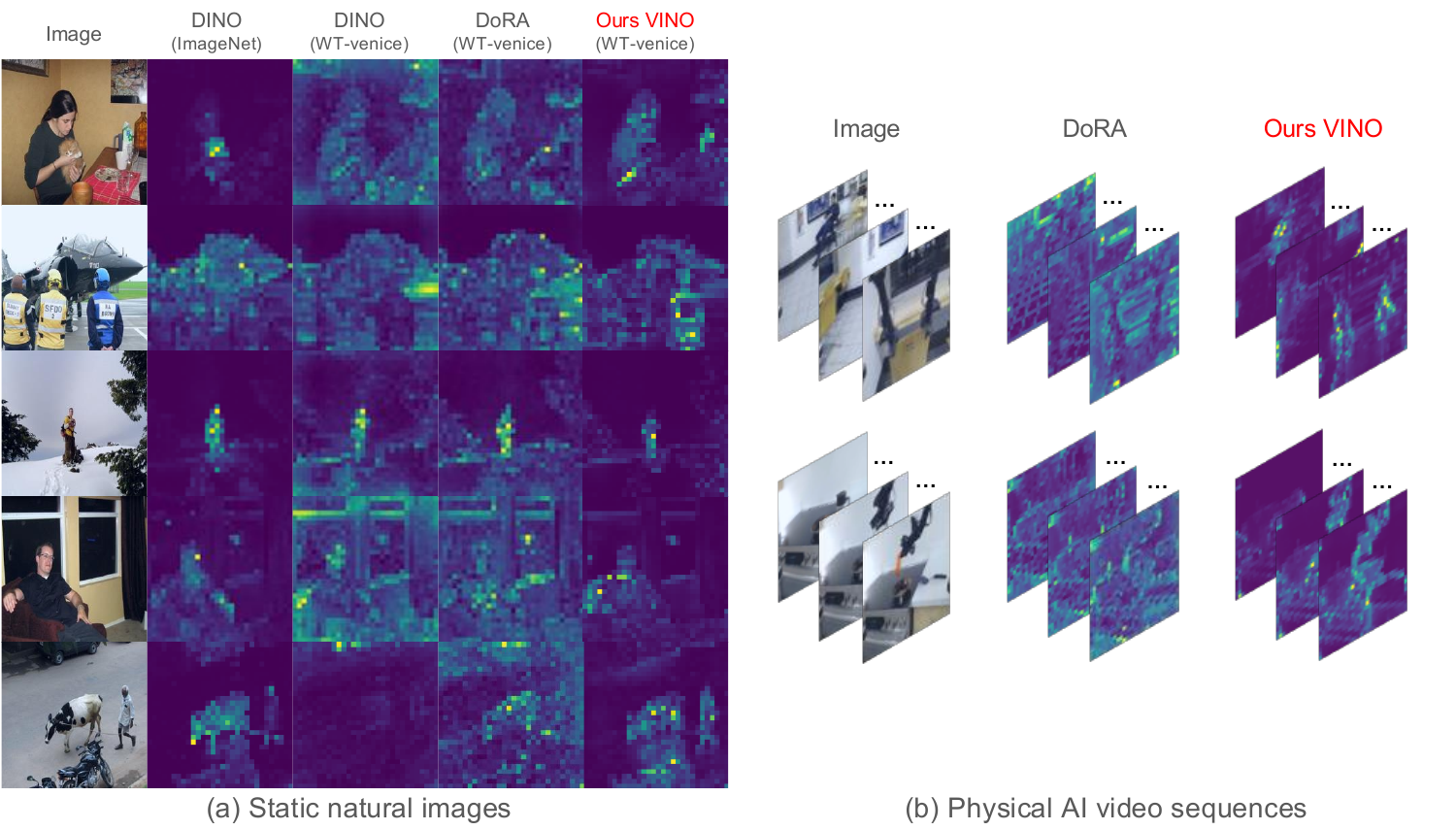}
  \caption{\textbf{Attention maps from ViT-S/16 encoders.} We visualize attention maps by ViT-S/16 encoders for the same inputs, comparing DINO trained on ImageNet, DINO trained on WT-Venice, DoRA, and VINO.
  \textbf{(a)} shows results on single static natural images from PASCAL VOC 2012\cite{Everingham2012PascalVOC2012}. \textbf{(b)} shows results on \emph{Physical AI} video sequences from Mobile ALOHA~\cite{Fu24MobileALOHA} dataset, where attention is visualized across multiple frames within each sequence.}
\label{fig:attn_map}
\end{figure}

\section{Introduction}
\label{sec:intro}
\paragraph{The Limits of Scaling Static Image SSL.} 
Self-supervised learning (SSL) has reached unprecedented levels of performance by pairing strong discriminative objectives with large-scale pretraining on curated image corpora~\cite{grill2020byol, caron2021dino, he2022mae}. A clear trend has emerged: robustness and transferability often improve with extreme spatial diversity and increasingly sophisticated data and training pipelines at web scale. For example, \textsc{DINOv2} relies on a curated collection of 142M images~\cite{oquab2024dinov2}, while \textsc{DINOv3} further pushes this scaling paradigm by using a curated subset of 1.6B images filtered from an initial pool of approximately 17B web-images~\cite{simeoni2025dinov3}. Concurrently, recent work shows that language-free visual SSL continues to scale favorably with data and model capacity, strengthening the case for pure-vision pretraining at web scale~\cite{fan2025webssl}. While successful, this trajectory is increasingly expensive---not only computationally, but also in terms of the data engineering and curation complexity required to build and maintain curated corpora. More importantly, scale and curation can statistically reduce reliance on contextual shortcuts by exposing objects under diverse contexts, but do not explicitly enforce figure--ground separation; this limitation becomes apparent on scene-centric data where object--background entanglement harms object-centric transfer~\cite{xie2021orl, li2022univip, wen2022slotcon}.

\begin{figure}[tb]
  \centering
  \includegraphics[width=1.0\columnwidth]{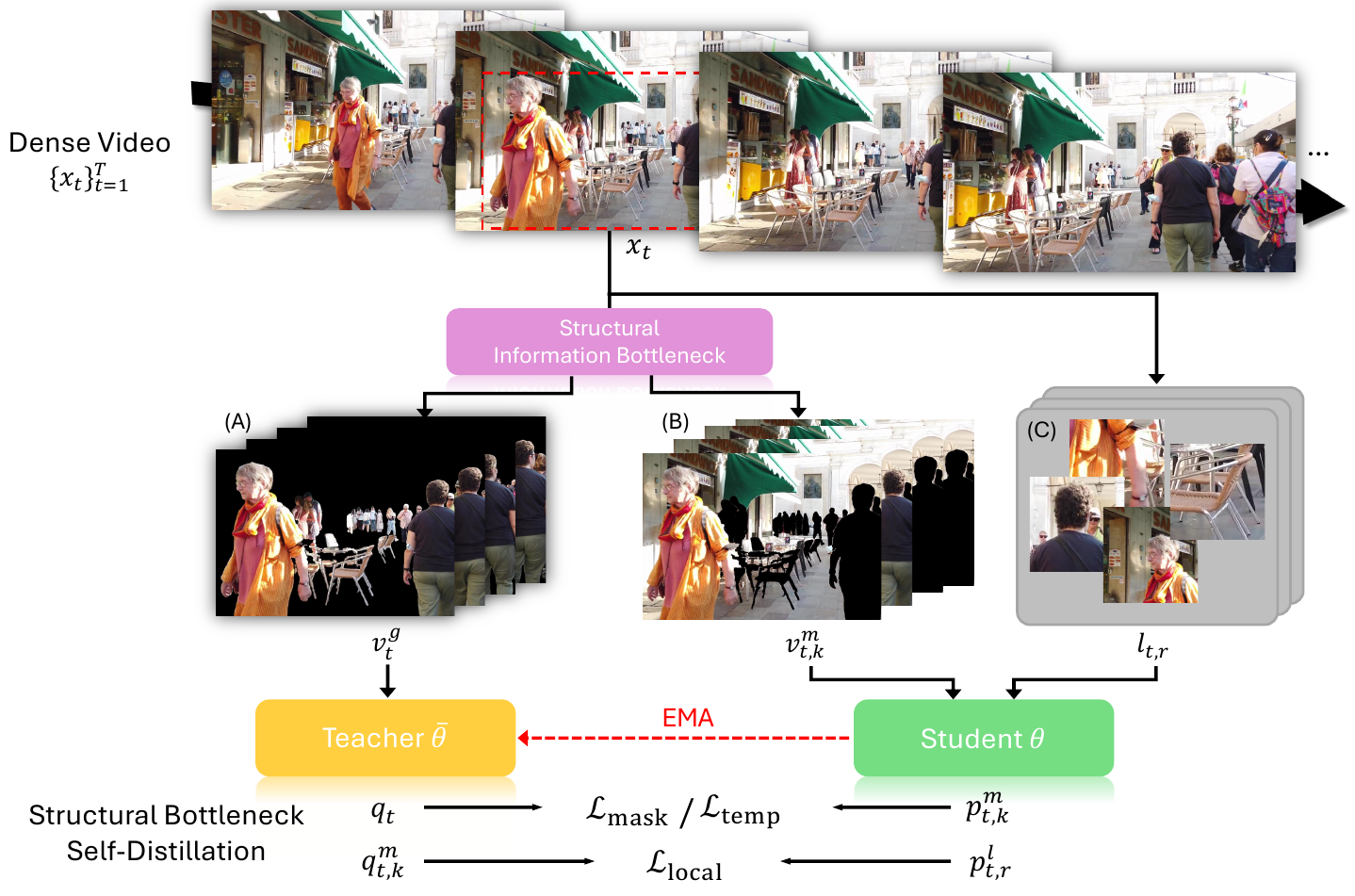}
  \caption{Our framework learns object-centric representations from dense video by enforcing a \textbf{structural information bottleneck}. 
(A) The \textbf{Teacher} observes a \emph{foreground-union} global view where background is suppressed, providing a de-contextualized target.
(B) The \textbf{Student} receives \emph{object-conditioned} views that retain background but remove co-occurring objects using a structural prior.
(C) This \textbf{asymmetric distillation} makes background and co-occurrence shortcuts non-predictive, pushing representations toward object-intrinsic cues while retaining robustness to natural context.
The total objective $\mathcal{L}_{\text{total}}$ ensures spatial de-contextualization ($\mathcal{L}_{\text{mask}}$), temporal object permanence ($\mathcal{L}_{\text{temp}}$), and part-to-whole consistency ($\mathcal{L}_{\text{local}}$).
  }
  \label{fig:overview}
\end{figure}
\paragraph{The Promise of Video Pretraining.} 
This motivates a complementary direction: learning from raw, \emph{in-the-wild} video streams, which are abundant and require minimal manual curation. Unlike static snapshots, video provides a continuous substrate of naturally occurring, identity-preserving transformations---viewpoint changes, deformations, interactions, and occlusions---that are difficult to synthesize with static crops alone. Temporal correspondence and cycle-consistency objectives exploit this structure to learn representations that remain consistent under time-varying appearance, offering a principled route to learning object permanence~\cite{dwibedi2019tcc, wang2019cycle, liu2025tcore}. Recent evidence further suggests that even a single long, uncurated video can become competitive with curated image benchmarks for learning transferable image encoders~\cite{VenkataramananR24DoRA}. Complementarily, large-scale object-centric video resources (\eg, millions of object tracks capturing evolving states) highlight the richness of object dynamics as self-supervision~\cite{wei2025trackverse}. However, learning from streaming video introduces orthogonal challenges: consecutive frames violate independent and identically distributed (IID) assumptions and can degrade representation learning if temporal correlation is not managed carefully~\cite{han2025orthograd}. To make this concrete, we summarize key structural properties of common pretraining corpora in Table~\ref{tab:dataset_properties_compact}. Notably, dense ego-motion videos (\eg, Walking Tours~\cite{VenkataramananR24DoRA}, BDD-100K~\cite{bdd100k}) exhibit high foreground sparsity and strong temporal coupling, a regime where temporal predictability can amplify contextual shortcuts rather than learning object permanence. This motivates objectives that explicitly control the information pathway from background context, \eg, by constructing de-contextualized \emph{targets} while keeping the \emph{inputs} context-rich for robustness.

\paragraph{The Co-occurrence Trap.} 
In dense naturalistic videos, this challenge manifests as the \textbf{Co-occurrence Trap}. In cluttered street scenes such as the Walking Tours video dataset~\cite{VenkataramananR24DoRA}, strong camera ego-motion couples foreground entities with persistent background structures. From the perspective of an SSL objective that rewards temporal predictability, the surrounding scene context (\eg, facades, pavements) becomes an unusually stable and accessible signal~\cite{ding2022fame, wang2021be}. Consequently, models can fall into \textbf{contextual overfitting}, learning representations that encode the surrounding environment rather than intrinsic object features; such context reliance is known to cause brittleness under background changes and weak transfer, and is particularly harmful for object-centric dense downstream tasks (\eg, detection and segmentation)~\cite{wang2021be, ding2022fame, mo2021objectaware, xie2021orl, li2022univip, wen2022slotcon, wang2021densecl, xie2021pixpro}. Existing dense-video SSL methods attempt to break this trap using internal guidance, such as attention-based tracks~\cite{VenkataramananR24DoRA} or motion priors from optical flow~\cite{wang2025poodle, teed2020raft}. We argue that these signals act as weak internal cues in ego-motion-heavy scenes: attention can drift toward persistent high-contrast background textures, and optical flow often reflects global camera motion more than local object dynamics. Without an explicit mechanism to enforce figure--ground separation, these methods struggle to disentangle objects from their surroundings, allowing contextual distractors to dominate the learned representation. This contextual entanglement is a significant hurdle for Physical AI. Embodied foundation models, such as \textsc{OpenVLA}~\cite{kim2024openvla}, increasingly rely on pretrained vision backbones for spatial grounding; yet, without explicit figure--ground separation, these agents remain vulnerable to ``visual distractors'' in unstructured environments~\cite{stone2021distracting}. Furthermore, while world models like \textsc{COSMOS}~\cite{nvidia2025cosmos} attempt to simulate physical reality through video generation, they require latent spaces that can explicitly disentangle the ``actor'' from the ``stage'' to learn true causality rather than mere background correlations. \textsc{VINO} addresses this gap by providing a scalable path to learn de-contextualized image encoders directly from raw video streams.

\begin{table}[t]
\centering
\newcommand{\tc}[1]{\begin{tabular}[c]{@{}c@{}}#1\end{tabular}}
\newcommand{\tl}[1]{\begin{tabular}[c]{@{}l@{}}#1\end{tabular}}

\setlength{\tabcolsep}{3pt}
\renewcommand{\arraystretch}{1.1}

\resizebox{\textwidth}{!}{%
\begin{tabular}{@{} l c c c c c c l @{}} 
\toprule
\textbf{Dataset} & \textbf{Modality} & \textbf{Curation} & \tc{\textbf{Spatial} \\ \textbf{Diversity}} & \tc{\textbf{FG} \\ \textbf{Sparsity}} & \tc{\textbf{Ego-} \\ \textbf{motion}} & \tc{\textbf{Temporal} \\ \textbf{Coupling}} & \tl{\textbf{Shortcut} \\ \textbf{Risk}} \\
\midrule
ImageNet~\cite{Deng09ImageNet} & Image & Curated & Medium & Low & None & None & Dataset bias \\
LVD-142M~\cite{oquab2024dinov2} & Image & \tc{Curated \\ (Web)} & \tc{Very \\ High} & \tc{Low--\\ Med} & None & None & \tl{Statistical \\ shortcut} \\
LVD-1689M~\cite{simeoni2025dinov3} & Image & \tc{Curated \\ (Billion)} & \tc{Extremely \\ High} & \tc{Low--\\ Med} & None & None & \tl{Statistical \\ shortcut} \\
Kinetics-400~\cite{kinetic}& Video & \tc{Semi-\\ curated} & Medium & Medium & Moderate & Medium & Scene bias \\
EPIC-Kitchens~\cite{Damen2018epickitchens} & Video & Egocentric & Medium & Medium & High & High & \tl{Motion \\ ambiguity} \\
BDD-100K~\cite{bdd100k} & Video & In-the-wild & Low & High & High & \tc{Very \\ High} & \tl{Structural \\ entanglement} \\
Walking Tours~\cite{VenkataramananR24DoRA} & Video & In-the-wild & \tc{Very \\ Low} & \tc{Very \\ High} & High & \tc{Extremely \\ High} & \tl{\textbf{Co-occurrence} \\ \textbf{trap}} \\
\bottomrule
\end{tabular}%
}
\vspace{2pt}
\caption{\textbf{Structural properties of pretraining corpora.}
Static curated image corpora (\eg, \textsc{DINOv2}/\textsc{DINOv3}) can dilute object--context correlations statistically via large spatial diversity,
whereas dense ego-motion videos exhibit strong temporal coupling and foreground sparsity, turning temporal predictability into contextual shortcuts.}
\label{tab:dataset_properties_compact}
\end{table}

\paragraph{Proposed Approach: \textsc{VINO}.} 
In this paper, we introduce \textsc{VINO} (\textit{Video-driven Invariance for Non-contextual Objects}), a framework designed to learn object-centric \emph{image encoders} from dense videos by enforcing a \textbf{structural information bottleneck}. Our key idea is to inject a class-agnostic structural prior as training scaffolding---not as pseudo-label supervision---to explicitly \emph{restrict the context pathway} during pretraining~\cite{kirillov2023sam, ravi2025sam2}. We implement an inverted asymmetric masked distillation task: the teacher is restricted to de-contextualized foreground-only views, providing a pure object-centric target, while the student processes context-rich views where background is preserved but competing object distractors are removed (\ie, inverted masking). This setup turns de-contextualization into a direct filtering requirement: to match the teacher's background-free representation, the student must learn to actively suppress the contextual noise present in its input and isolate object-intrinsic features. To retain part-to-whole generalization without degenerating into background texture matching in multi-object scenes, \textsc{VINO} additionally uses \emph{object-aware local views} sampled around foreground regions.

Furthermore, to exploit temporal richness without background leakage, \textsc{VINO} introduces teacher-anchored cross-time distillation over short temporal tubes (\eg, $T{=}4$ frames). Using track-consistent object identities within the tube, we align the teacher's pure foreground representation at time $t_i$ with the student's context-aware masked representation at time $t_j$. This cross-time distillation enforces temporal object permanence, requiring the student to extract the same invariant object features regardless of viewpoint shifts, deformations, or the presence of background context, all while maintaining training stability. Crucially, because the teacher targets are de-contextualized by construction, this cross-time consistency reinforces object identity rather than scene-level predictability~\cite{liu2025tcore}.

\paragraph{Contributions.}
Our contributions are three-fold:
\begin{itemize}
    \item \textbf{Formalizing the Co-occurrence Trap:} We identify and formalize the co-occurrence trap in dense ego-motion videos, explaining why temporal predictability drives contextual overfitting in existing attention- or motion-guided SSL approaches.
    \item \textbf{Structural Information Bottleneck:} We propose \textsc{VINO}, introducing an \textit{inverted structural information bottleneck}. By distilling from a foreground-only teacher to a context-aware student, we force the student encoder to learn active background suppression, making de-contextualization a primary optimization goal.
    \item \textbf{Unsupervised Object Discovery:} We demonstrate that \textsc{VINO} learns shape-biased and object-centric representations possessing intrinsic figure-ground separation capabilities. Through attention visualization and unsupervised object discovery, we show that \textsc{VINO} significantly surpasses recent dense-video SSL methods in isolating objects from complex backgrounds (see Sec.~\ref{sec:experiments}).
\end{itemize}

\section{Related Work}
\label{sec:relatedwork}
\subsection{Scaling Image SSL and the Persistence of Contextual Shortcuts}
Self-supervised learning for vision has progressed from instance discrimination and contrastive learning~\cite{chen2020simclr, he2020moco, caron2020swav} to non-contrastive/predictive objectives and masked modeling~\cite{grill2020byol, chen2021simsiam, caron2021dino, he2022mae, zbontar2021barlowtwins, bardes2022vicreg}. Recent ``foundation'' encoders scale these objectives with web-scale curation and filtering, producing highly transferable features~\cite{oquab2024dinov2, simeoni2025dinov3, fan2025webssl}. Yet a recurring concern is that strong SSL objectives can exploit spurious correlations and contextual shortcuts when the data distribution systematically couples objects and backgrounds, harming robustness and transfer in settings where context shifts~\cite{park2024lowrank, kim2024keywordbias, hamidieh2024featurespaceaug}. A parallel line of work attempts to make SSL more object-aware through region-level matching, grouping, or object-level bootstrapping~\cite{xie2021orl, wen2022slotcon, li2022univip, lebailly2024cribo, kakogeorgiou2024spot, djukic2025ocebo}. These approaches support the broader thesis that how the learning signal is grounded (global scene vs.\ object-centric factors) matters as much as data scale. \textsc{VINO} builds on this perspective, but targets a regime where dataset-level ``dilution'' of context (via massive static-image diversity) is unavailable: dense, uncurated videos where context and object co-occur persistently.

\subsection{Video as Self-Supervision: Temporal Correspondence, Streaming, and Dense Natural Videos}
Video SSL leverages temporal coherence as a natural supervisory signal. Early pretext tasks used temporal ordering and prediction~\cite{misra2016shuffle, xu2019vcop}, while correspondence-based methods learn embeddings stable to viewpoint, deformation, and occlusion by enforcing cycle-consistency or temporal alignment~\cite{dwibedi2019tcc, wang2019cycle}. More recent methods incorporate temporal correspondence into masked modeling or distillation frameworks to better handle temporal uncertainty and learn higher-level semantics~\cite{liu2025tcore}. At the same time, learning directly from streaming video raises non-IID optimization issues and ``one-to-many'' target ambiguity, motivating recent remedies that explicitly address correlated gradients or conditional uncertainty in SSL pairs~\cite{han2025orthograd}. Closest to our setting, dense \emph{in-the-wild} video pretraining has been explored through long, uncurated streams~\cite{VenkataramananR24DoRA} and through object-centric track datasets that expose state and pose dynamics~\cite{wei2025trackverse}. However, dense ego-motion videos pose a distinctive failure mode: the temporal signal can be dominated by persistent scene layout, making background correspondence as ``predictable'' as the foreground. \textsc{VINO} is positioned at this junction: it aims to retain the benefits of temporal supervision while preventing temporal coherence from collapsing into scene/context encoding.

\subsection{Inductive Priors for Object Discovery in Dense Videos}
To prevent background leakage, recent dense-video SSL methods introduce inductive cues to localize or track foreground content without labels. \textsc{DoRA} adopts attention-based tracks from the model itself~\cite{VenkataramananR24DoRA}, while \textsc{PooDLe} adopts motion priors derived from optical flow~\cite{wang2025poodle, teed2020raft}. These cues can be effective in moderately structured videos but become unreliable in ego-motion-heavy scenes: attention can drift toward salient background textures, and flow is often dominated by global camera motion rather than object-specific dynamics. In parallel, the emergence of strong class-agnostic segmentation and structural models provides an alternative form of guidance—high-fidelity figure--ground information that is less entangled with motion magnitude or early-training saliency~\cite{kirillov2023sam, ravi2025sam2}. Complementing this trend, recent evidence suggests that injecting perceptual/structural inductive bias (\eg, silhouette- or figure--ground-based bias) can be a prerequisite for robust representation learning in cluttered real imagery~\cite{zhao2025perceptualbias}. \textsc{VINO} follows this direction, but uses structural cues not as pseudo-label supervision; instead, it uses them as \emph{training scaffolding} to impose a structural information bottleneck within self-distillation. This design explicitly restricts the student’s access to context and forces the teacher to suppress background dependencies, so that temporal correspondence is learned over de-contextualized object representations rather than over predictable scene layout.

\section{Method}
\label{sec:method}
\subsection{Training Setup and View Generation}
\textsc{VINO} trains an image encoder from scratch using dense, uncurated videos as a source of naturally correlated frames.
Given a video, we randomly sample a short tube of $T$ frames $\{x_t\}_{t=1}^{T}$ with a fixed time step.
Each frame is encoded independently; temporal structure is used only to define cross-time correspondences. To ensure spatial alignment across time, we apply shared spatial transforms to all frames in the tube. Specifically, we first sample a scene-level pre-crop shared across all frames $t$ in the tube, followed by a global crop. The resulting global crop $g_l$ defines a shared scene window across time, ensuring spatial alignment. We use this common geometry to generate both teacher and student inputs by combining standard photometric augmentations with class-agnostic structural masking.

A class-agnostic structural prior provides per-pixel instance masks for each frame. Applying the same spatial transforms to the masks yields object masks in the global-crop coordinate system:
\begin{equation}
\{m_{t,k}\}_{k=1}^{K_t}, \qquad m_{t,k}\in\{0,1\}^{H\times W}, 
\end{equation}
where $K_t$ varies across frames and we keep up to $M$ objects per frame (\eg, $M{=}10$) after confidence/size filtering.
Crucially, the mask pipeline also provides tube-level \emph{track-consistent} identities $\mathrm{id}(t,k)$, enabling temporal matching within the sampled tube (please note that details of mask generation/association are implementation-specific and do not involve semantic labels).

\subsubsection{Teacher foreground-union view and student object-conditioned views.}
Let $\phi(\cdot)$ denote photometric augmentation and resizing applied to the global crop.
We first form a context-preserving base view
\begin{equation}
v_t^{\mathrm{full}} = \phi(g_t).
\end{equation}
Let $m_t^{\cup}=\bigvee_{k=1}^{K_t}m_{t,k}$ be the resized union foreground mask. We suppress background by keeping only the union of foreground instances:
\begin{equation}
v_t^{g} = v_t^{\mathrm{full}} \odot m_t^{\cup}.
\end{equation}
For the student masked view (object-conditioned scene) corresponding to object $k$, we keep the selected object \emph{and} the background while suppressing all other co-occurring objects:
\begin{equation}
v_{t,k}^{m} = v_t^{\mathrm{full}}\odot \tilde{m}_{t,k},
\qquad
\tilde{m}_{t,k} = (1-m_t^{\cup}) + m_{t,k}.
\label{eq:masked-view}
\end{equation}
Thus, teacher targets are computed from a background-suppressed foreground-union view,
whereas the student observes an object-conditioned view that removes co-occurring objects under identical crop geometry.

\subsubsection{Mask-guided local views}
Local-to-global distillation is beneficial for part-to-whole generalization, but in cluttered multi-object scenes naive random locals may share little semantics with the teacher global view.
We therefore sample local crops conditioned on foreground overlap.
Let $m_t^{\cup}$ be the union foreground mask in the \emph{global-crop} coordinate system.
We sample $R$ local crops $\{\ell_{t,r}\}_{r=1}^{R}$ as \emph{sub-crops of the same global window} such that each crop overlaps sufficiently with $m_t^{\cup}$:
\begin{equation}
\frac{|\mathcal{R}(\ell_{t,r}) \cap m_t^{\cup}|}{|\mathcal{R}(\ell_{t,r})|} \ge \alpha,
\label{eq:local-overlap}
\end{equation}
where $\mathcal{R}(\cdot)$ denotes the crop region and $\alpha$ is a fixed threshold.
This ensures local views capture meaningful object parts (rather than arbitrary background patches), making local-to-global alignment well-posed in dense videos.

\subsection{Structural Bottleneck Self-Distillation Objective}
We adopt an EMA teacher--student self-distillation framework.
The student $\theta$ processes object-conditioned masked views $v_{t,k}^{m}$ and local views $\ell_{t,r}$,
while the teacher $\bar{\theta}$ processes only the background-suppressed foreground-union global views $v_t^{g}$.
The teacher is updated as an exponential moving average of the student:
\begin{equation}
\bar{\theta} \leftarrow \mu \bar{\theta} + (1-\mu)\theta.
\end{equation}

Let $z_{\theta}(\cdot)$ and $z_{\bar{\theta}}(\cdot)$ denote the projected logits (backbone + head).
Following \textsc{DINO}-style self-distillation, we form categorical distributions with temperatures and a running center $c$:
\begin{align}
q_t &= \mathrm{Softmax}\!\left(\frac{z_{\bar{\theta}}(v_t^{g}) - c}{\tau_T}\right), \\
p_{t,k}^{m} &= \mathrm{Softmax}\!\left(\frac{z_{\theta}(v_{t,k}^{m})}{\tau_S}\right), \\
p_{t,r}^{\ell} &= \mathrm{Softmax}\!\left(\frac{z_{\theta}(\ell_{t,r})}{\tau_S}\right).
\end{align}
The basic distillation term is the cross-entropy
\begin{equation}
H(q,p) = -\sum_{u} q(u)\log p(u),
\end{equation}
which is equivalent to minimizing $\mathrm{KL}(q\|p)$ w.r.t.\ the student since $H(q,p)=H(q)+\mathrm{KL}(q\|p)$ and $q$ is not updated by gradient descent.

\subsubsection{Structural Information Bottleneck (SIB)}
Our key mechanism is an \emph{operational} information bottleneck induced by structural masking in the teacher and student pathways.
Teacher targets are computed from $v_t^{g}$, a foreground-union view where background pixels are suppressed,
making the teacher distribution invariant to background by construction.
Meanwhile, student masked inputs $v_{t,k}^{m}$ remove \emph{co-occurring objects} while retaining the selected object and surrounding background,
weakening co-occurrence shortcuts in dense multi-object videos.
Under distillation, the student is trained to match background-suppressed teacher targets despite observing natural context,
biasing the learned representation toward object-intrinsic cues.
Importantly, masks are used only to control information pathways (not as semantic pseudo-label supervision).

\paragraph{Spatial de-contextualization via masked distillation.}
We align each masked object view to the teacher global distribution from the same frame:
\begin{equation}
\mathcal{L}_{\mathrm{mask}}
=
\frac{1}{TK_t}\sum_{t=1}^{T}\sum_{k=1}^{K_t}
H\!\left(q_t,\; p_{t,k}^{m}\right).
\label{eq:mask-loss}
\end{equation}
Because teacher targets $q_t$ are computed from background-suppressed foreground-union views and student masked inputs remove co-occurring objects, minimizing~\eqref{eq:mask-loss} discourages representations that rely on background or co-occurrence cues, thereby mitigating contextual shortcuts.

\paragraph{Temporal object permanence via teacher-anchored cross-time distillation.}
To exploit temporal coherence while maintaining the structural bottleneck, we impose cross-time consistency
between teacher targets computed from foreground-union views and student object-conditioned masked views.
Using track-consistent identities $\mathrm{id}(t,k)$, we define cross-time positives within the tube:
\begin{equation}
\mathcal{P}=
\left\{
(t,k,t')\;\middle|\;
t'\neq t,\ \exists k' \text{ s.t. } \mathrm{id}(t,k)=\mathrm{id}(t',k')
\right\}.
\end{equation}
We then distill the teacher's global distribution at time $t'$ into the student's masked distribution at $(t,k)$:
\begin{equation}
\mathcal{L}_{\mathrm{temp}}
=
\frac{1}{|\mathcal{P}|}\sum_{(t,k,t')\in\mathcal{P}}
H\!\left(q_{t'},\; p_{t,k}^{m}\right).
\label{eq:temp-loss}
\end{equation}
Anchoring to the EMA teacher preserves training stability while encouraging identity-consistent representations across viewpoint/deformation/occlusion changes.

\paragraph{Part-to-whole generalization via mask-guided local-to-global distillation.}
Finally, we distill the teacher global distribution to mask-guided local views:
\begin{equation}
\mathcal{L}_{\mathrm{local}}
=
\frac{1}{TR}\sum_{t=1}^{T}\sum_{r=1}^{R}
H\!\left(q_t,\; p_{t,r}^{\ell}\right).
\label{eq:local-loss}
\end{equation}
Since the teacher target is computed from a background-suppressed view, $\mathcal{L}_{\mathrm{local}}$ further discourages
degenerate alignment via background textures and promotes part-to-whole consistency. Final training objective is
\begin{equation}
\mathcal{L}
=
\lambda_{\mathrm{local}}\mathcal{L}_{\mathrm{local}}
+\lambda_{\mathrm{mask}}\mathcal{L}_{\mathrm{mask}}
+\lambda_{\mathrm{temp}}\mathcal{L}_{\mathrm{temp}}.
\label{eq:total}
\end{equation}
In practice, $\mathcal{L}_{\mathrm{mask}}$ and $\mathcal{L}_{\mathrm{temp}}$ are computed over valid masks/IDs only, while $\mathcal{L}_{\mathrm{local}}$ remains active due to mask-guided sampling.
Together, the masked bottleneck first suppresses contextual shortcuts, and the cross-time constraint then converts temporal coherence into robust object-centric invariances.

\section{Experiments}
\label{sec:experiments}

\subsection{Experimental Setups}
\label{sec:exp_setup}

\subsubsection{Implementation Details}
Given a video, we randomly sample a short tube of $T=4$ frames with a fixed stride of \num{10}.
The softmax temperatures are set to $\tau_\mathrm{S} = 0.1$ and $\tau_\mathrm{T} = 0.04$.
The batch size is \num{768} realized by gradient accumulation. 
The loss weights are $\lambda_{\mathrm{local}} = 1.0$, $\lambda_{\mathrm{mask}} = 0.5$, $\lambda_{\mathrm{temp}} = 0.5$ and set heuristically.
The pre-crop is the image region that includes all foreground objects.
Local crops are taken randomly from foreground objects with a relative area of $(0.4, 0.7)$ taken from the object's bounding box, which was padded by $10\%$ in each direction.
Object masks are pre-computed and produced by \textsc{SAM3}~\cite{carion2025sam3segmentconcepts}.
In the rare case if no objects are present in a single frame, we only use local crops extracted from the whole image according to the strategy from \textsc{DINO}.

\subsubsection{Baselines}
\label{sec:exp_baselines}
We compare \textsc{VINO} with recent state-of-the-art self-supervised learning methods, including \textsc{iBOT}~\cite{Zhou22iBOT}, \textsc{AttMask}~\cite{Kakogeorgiou22AttMask}, \textsc{DINO}~\cite{caron2021dino}, \textsc{DINOv2}~\cite{oquab2024dinov2}, and \textsc{DoRA}~\cite{VenkataramananR24DoRA}. 
For the fair comparison, we introduce the same image encoder, ViT-S with patch size of 16 and identical downstream pipelines across methods, changing only the self-supervised initialization of the image encoders.

\subsubsection{Datasets}
\label{sec:exp_dataset}
We train \textsc{VINO} in a fully self-supervised manner using a single long-form video from the Venice sequence of the Walking Tours~\cite{VenkataramananR24DoRA}.
The raw footage is captured in a dynamic urban environment with continuous camera motion, inducing naturally occurring appearance changes (\eg, viewpoint and scale shifts, partial occlusions, and illumination transitions).
We exploit this temporal continuity as an implicit supervisory signal to learn invariances without manual curation.

The original WT-Venice video is recorded at $1920\times1080$ resolution and $60$ FPS with a total duration of $1\,\mathrm{h}\,50\,\mathrm{min}$, which corresponds to approximately 400K frames. For view generation, we adopt multi-crop augmentation and apply the transformations to both global and local crops, including color jitter, Gaussian blur, and solarization~\cite{caron2021dino}.
By relying on a single densely informative video rather than curated object-centric image collections such as ImageNet-1K~\cite{Deng09ImageNet}, our setup probes whether robust and transferable representations can emerge from a continuous visual experience in the real-world alone.

\subsubsection{Evaluated Tasks: Probing Intrinsic Structure}
\label{sec:exp_downstream}
To assess whether our structural bottleneck successfully mitigates the co-occurrence trap, we focus on evaluations that directly probe the intrinsic structure of the learned features, avoiding heavy downstream fine-tuning which might mask the quality of the raw representation.

\paragraph{Unsupervised Object Discovery.}
We evaluate unsupervised single-object localization on PASCAL VOC 2012~\cite{Everingham2012PascalVOC2012} (trainval set) using the LOST method~\cite{Simeoni21LOST}.
LOST operates on a single image without external supervision. It relies on the observation that background patches in self-supervised transformers tend to be highly correlated with each other, whereas foreground patches are less correlated with the background. 
Methodologically, we feed the image to the frozen VINO encoder and extract the keys ($k$) from the last self-attention layer. LOST computes a patch similarity graph, selects a ``seed'' patch with the lowest degree (inverse density), and expands it to correlated neighbors to form a detection box. 
We report the \textbf{CorLoc} metric (Correct Localization), defined as the percentage of images where the intersection-over-union (IoU) between the predicted box and one of the ground-truth bounding boxes is $\ge 0.5$. This metric directly tests if the encoder's feature correlations respect figure--ground boundaries.

\begin{figure}[t!]
  \centering
  \includegraphics[width=1.0\columnwidth]{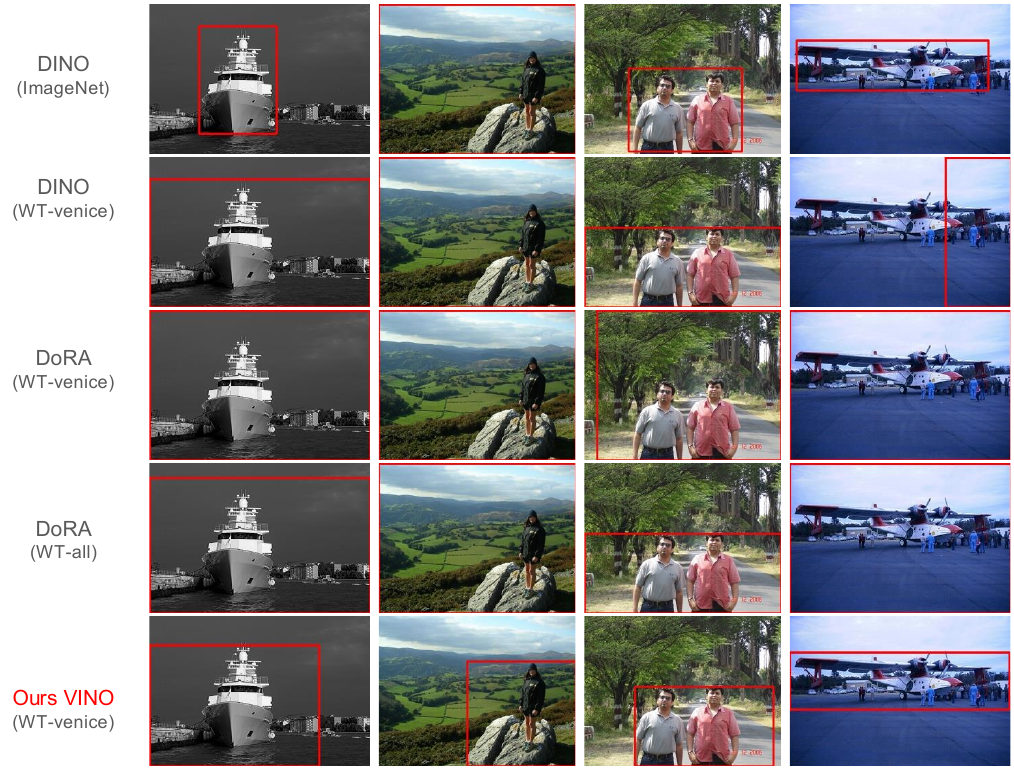}
  \caption{\textbf{Unsupervised object discovery on PASCAL VOC 2012.}
  We visualize the predicted object bounding boxes obtained from attention based foreground masks following the default LOST\cite{Simeoni21LOST}. We compared results on ViT-S/16 encoders for the same inputs, comparing DINO trained on ImageNet, DINO trained on WT-Venice, DoRA, and VINO.
  Compared to baselines, \textsc{VINO} produces tighter boxes that better align with the principal object extent and is less prone to drifting toward large background regions, highlighting improved figure--ground separation under dense ego-motion pretraining.}
  \label{fig:voc_discovery_boxes}
\label{fig:lost_pred}
\end{figure}

\section{Experimental Results}
We evaluate whether \textsc{VINO}'s structural bottleneck translates into improved object-centricity from a single dense ego-motion video.
Walking Tours--Venice exhibits strong temporal coupling and heavy object--background co-occurrence.
In this regime, improvements should appear most clearly in the model's ability to distinguish figure from ground. We analyze this qualitatively via attention maps and quantitatively via unsupervised object discovery.

\subsection{Qualitative Analysis: Attention Visualizations}

\paragraph{Emergent Attention Properties.}
We first examine the emergent attention properties of \textsc{VINO}. Fig.~\ref{fig:attn_map} visualizes the self-attention maps of the [CLS] token from the last layer of ViT-S/16 encoders. We compare \textsc{VINO} against \textsc{DINO} (trained on ImageNet and WT-Venice) and \textsc{DoRA}. We visualize the attention using images from PASCAL VOC 2012\cite{Everingham2012PascalVOC2012} (Fig.~\ref{fig:attn_map}\textbf{a}). To investigate the transferability of the backbone trained on WT-Venice to application tasks, we also use \emph{Physical AI} manipulation video sequences from Mobile ALOHA\cite{Fu24MobileALOHA} (Fig.~\ref{fig:attn_map}\textbf{b}; tasks: \emph{Push Chairs} and \emph{Cook Shrimp}) for visualization of the same backbone.

Fig.~\ref{fig:attn_map}\textbf{a} displays attention on static images from PASCAL VOC 2012. Baselines trained on dense video (\textsc{DINO} on WT-Venice, \textsc{DoRA}) often exhibit "leakage," where attention spreads to high-contrast background textures or covers the entire scene, reflecting the contextual overfitting inherent to the data source. In contrast, \textsc{VINO} produces sharp, shape-aligned attention maps that tightly encompass the foreground object. This suggests that the structural information bottleneck successfully forces the model to discard background correlations during pretraining.

\paragraph{Transfer to Physical AI.}
As a qualitative evaluations, Fig.\ref{fig:attn_map} visualizes attention maps extracted from ViT-S/16 encoders for the same inputs across different pretrained backbones.
On both single static images from PASCAL VOC 2012\cite{Everingham2012PascalVOC2012} (Fig.~\ref{fig:attn_map}\textbf{a}) and \emph{Physical AI} manipulation video sequences from Mobile ALOHA\cite{Fu24MobileALOHA} (Fig.~\ref{fig:attn_map}\textbf{b}; tasks: \emph{Push Chairs} and \emph{Cook Shrimp}), VINO shows attention that is more concentrated on salient foreground regions, whereas baselines---particularly those trained on dense ego-motion video---exhibit diffuse responses that extend to persistent background textures. 
Beyond standard vision benchmarks, (Fig.~\ref{fig:attn_map}\textbf{b}) provides a qualitative results in a \emph{Physical AI} setting. Embodied manipulation is inherently scene-centric: the robot body, workspace geometry, and repetitive background structures remain highly persistent across time, making strong opportunities for context shortcuts.
Despite this, VINO maintains object-aligned attention over multiple frames, indicating that the learned features prioritize task-relevant entities (e.g., the manipulated chair and contact regions) rather than stable scene textures. This observation supports our main contribution that context-controlled distillation under a structural bottleneck yields representations that are less entangled with background cues, and thus better suited for object-centric transfer in cluttered, temporally correlated environments.

\paragraph{Unsupervised Object Discovery.}
In addition, Fig.~\ref{fig:voc_discovery_boxes} provides qualitative evidence for improved object-centric localization under our structural bottleneck.
Using the standard LOST evaluation pipeline~\cite{Simeoni21LOST}, we convert attention maps into foreground masks and extract tight bounding boxes without any supervision.
Across diverse VOC images, baselines trained on dense WT-Venice video often exhibit \emph{contextual drift}, producing oversized or misplaced boxes that extend to salient background structures.
In contrast, \textsc{VINO} yields more compact, object-aligned boxes that better capture the dominant object extent, consistent with reduced background leakage during pretraining.
This qualitative behavior complements the CorLoc improvements reported in Table~\ref{tab:downstream_discovery} and supports our methods that context-controlled distillation encourages figure--ground separation.

\subsection{Feature Recognition}
\subsubsection{Unsupervised Object Discovery.}
Table~\ref{tab:downstream_discovery} summarizes unsupervised object discovery on PASCAL VOC 2012~\cite{Everingham2012PascalVOC2012}. This task directly probes the spatial structure of the learned representation: a high CorLoc indicates that the model's internal attention naturally aligns with object extents rather than background context or texture.
Among baselines trained on WT-Venice, \textsc{iBOT} and \textsc{DoRA} perform competitively, achieving $33.9\%$ and $30.4\%$ CorLoc respectively. Conversely, \textsc{PooDLe} lags significantly ($22.6\%$), suggesting that motion-based grouping alone may fail to separate objects from the global flow in dense ego-motion videos.
\textsc{VINO} outperforms all baselines with a CorLoc of 34.8\%, surpassing the strongest competitor by +$\Delta$0.9. This substantial gain confirms that our structural bottleneck effectively suppresses the co-occurrence trap, forcing the encoder to discover and localize object-intrinsic features even without manual annotations.

\begin{table}[t]
\centering
\scriptsize
\caption{\textbf{Object discovery.} We report unsupervised object discovery performance on PASCAL VOC 2012~\cite{Everingham2012PascalVOC2012} measured by CorLoc. All methods use ViT-S/16 pretrained on the Walking Tours dataset.}
\label{tab:downstream_discovery}
\begin{tabular}{l c l c}
\toprule
\multirow{2}{*}{Method} & \multirow{2}{*}{Epochs} & \multirow{2}{*}{Pre-train} &
\multicolumn{1}{c}{Object Discovery} \\
\cmidrule(lr){4-4}
& & & $\mathrm{CorLoc}\uparrow$ \\
\midrule
\textsc{DoRA}~\cite{VenkataramananR24DoRA}     & 100 & WT-All    & 33.7 \\
\midrule
\textsc{PooDLe}~\cite{wang2025poodle}          & 100 & WT-Venice & 22.6 \\
\textsc{DINO}~\cite{caron2021dino}             & 100 & WT-Venice & 24.8 \\
\textsc{DINOv2}~\cite{oquab2024dinov2}         & 100 & WT-Venice & 27.5 \\
\textsc{iBOT}~\cite{Zhou22iBOT}                & 100 & WT-Venice & 33.9 \\
\textsc{DoRA}~\cite{VenkataramananR24DoRA}     & 100 & WT-Venice & 30.4 \\
\rowcolor{gray!15}
\textbf{\textsc{VINO} (ours)}                 & 100 & WT-Venice & \textbf{34.8}  \\
\bottomrule
\end{tabular}
\end{table}

\section{Conclusion}
In this work, we introduced \textsc{VINO}, a self-supervised learning framework designed to mitigate the contextual co-occurrence trap inherent to dense, ego-motion video pretraining. By imposing a structural information bottleneck via asymmetric masked distillation---where a context-aware student is forced to predict de-contextualized, foreground-only teacher targets---\textsc{VINO} actively discourages the representation from relying on persistent background shortcuts. 

Rather than relying on broad fine-tuned benchmarks that can obscure the raw quality of pretraining, we analyzed the intrinsic properties of the learned features directly. Our qualitative analysis of attention maps and quantitative evaluation on unsupervised object discovery demonstrate that \textsc{VINO} effectively decouples objects from their environment. This indicates that strategically controlling \emph{what a model learns to ignore} is a powerful mechanism for learning robust, object-centric representations from uncurated visual streams, paving the way for more efficient and disentangled perception in autonomous systems.


%
%
\bibliographystyle{splncs04}
\bibliography{main}

\end{document}